\documentclass[letterpaper]{article} 
\usepackage[preprint]{neurips_2021}  
\usepackage{booktabs}
\usepackage{times}  
\usepackage{helvet}  
\usepackage{courier}  
\usepackage{url}  
\usepackage{graphicx}  
\usepackage{rotating}
\usepackage{wrapfig}
\usepackage{multicol}
\usepackage{latexsym}
\usepackage{amsmath}
\usepackage{amssymb}
\usepackage{url}
\usepackage[noend]{algorithmic}
\usepackage{algorithm}
\usepackage{multirow}
\usepackage{mathtools}
\usepackage{enumitem}
\usepackage{color}
\usepackage{xcolor}         
\usepackage{microtype}
\usepackage{tabularx}
\usepackage{relsize}
\usepackage{amsthm}
\usepackage{subfigure}

\newtheorem{defn}{Definition}

\newcommand{\C}{\mathcal{C}}

\newcommand{\Org}{\textsf{Organization}}

\begin{document}

\title{Many-Agent Reinforcement Learning\\ under Partial Observability}  

\author{%
  Keyang He \\
  Department of Computer Science\\
  University of Georgia\\
  \texttt{keyang@uga.edu} \\
  \And
  Prashant Doshi \\
  Department of Computer Science\\
  University of Georgia\\
  \texttt{pdoshi@uga.edu} \\
  \AND
  Bikramjit Banerjee \\
  Department of Computer Science\\
  University of Southern Mississippi \\
  \texttt{Bikramjit.Banerjee@usm.edu} \\
}
\maketitle

\begin{abstract}  
Recent renewed interest in multi-agent reinforcement learning (MARL) has generated an impressive array of techniques that leverage deep reinforcement learning, primarily actor-critic architectures, and can be applied to a limited range of settings in terms of observability and communication. However, a continuing limitation of much of this work is the curse of dimensionality when it comes to representations based on joint actions, which grow exponentially with the number of agents. In this paper, we squarely focus on this challenge of scalability. We apply the key insight of {\em action anonymity}, which leads to permutation invariance of joint actions, to two recently presented deep MARL algorithms, MADDPG and IA2C, and compare these instantiations to another recent technique that leverages action anonymity, viz., mean-field MARL.  We show that our intantiations can learn the optimal behavior in a broader class of agent networks than the mean-field method, using a recently introduced pragmatic domain. 
\end{abstract}


\section{Introduction}
\label{sec:intro}

Continued interest in multi-agent reinforcement learning (MARL) has yielded a variety of algorithms over the years, from Minmax-Q~\cite{minimaxq} and Nash-Q~\cite{nashq} during its initial study to the more recent ones such as MADDPG~\cite{maddpg}, COMA~\cite{coma}, and IA2C~\cite{ia2c}. While the early methods mostly generalized Q-learning~\cite{qlearning} to multiagent settings, the later methods utilize the actor-critic schema with centralized or decentralized actor and critic components. The neural network representations of the actor and critic components allow these methods, which by default target settings with perfectly observed states, to expand to partial observability by maintaining a moving window of past observations. While these methods have demonstrated good performance on the standard MARL problem domains, the RL does not practically scale beyond a handful of interacting agents.  

Multiagent planning frameworks such as DEC- and I-POMDPs~\cite{decpomdp,ipomdp} faced a similar hurdle of scaling in a meaningful way to many agents. A key insight -- that many domains exhibit the {\em action anonymity} structure~\cite{Jovanovic88:Anonymous} -- helped mitigate this curse of many agents afflicting planning. More specifically, it is the number of agents that perform the various actions which matters to the reasoning rather than the respective identities of the agents performing the actions; in other words, joint action permutations are equivalent. Modeling this invariance enables the planning complexity to drop from being exponential in the number of agents to polynomial thereby facilitating multiagent planning for thousands of agents~\cite{Sonu17:Anonymous,Varakantham14:Decentralized}.

In this paper, we aim to bring this insight to MARL and scale the learning to many-agent settings. An existing MARL approach that implicitly presumes action anonymity (perhaps, without being aware of this presumption) suggests using the average action vector, obtained as the mean of one-hot encodings of actions of the other agents~\cite{MF}. {\em This mean-field value is a near-optimal approximation under action anonymity when the multiagent system can be decomposed into pairwise interactions.} Our first contribution is an elucidation of the approximation that the mean-field necessitates and, in response, a general technique that represents other agents' behavior under action anonymity using {\em action configurations} without loss in value. This new method does not require approximating agent interactions as pairwise ones.

We integrate both the mean-field approximation and the generic action configuration based representations in two recent MARL approaches: MADDPG and IA2C. Our second contribution is a comprehensive comparison of the scalable representations on a recently introduced cooperative-competitive \Org{} domain~\cite{ia2c} that can organically scale to many agents and exhibits action anonymity. We show that interaction topologies exist where the two representations yield identical values, which is also the optimal joint behavior. However, several topologies also exist where the learned behaviors and values differ, and that the mean-field converges to suboptimal policies. Finally, our third contribution is a demonstration of MARL under partial observability in settings that contain up to a hundred agents.

\section{Background}
\label{sec:background}

We briefly describe the domain for modeling an organization in the next subsection, and follow it up with a review of the recently introduced {\em interactive A2C} (IA2C) method for learning in mixed cooperative-competitive settings under partial observability.

\subsection{The Organization domain}
\label{subsec:org}

\begin{figure}
\includegraphics[width=\linewidth]{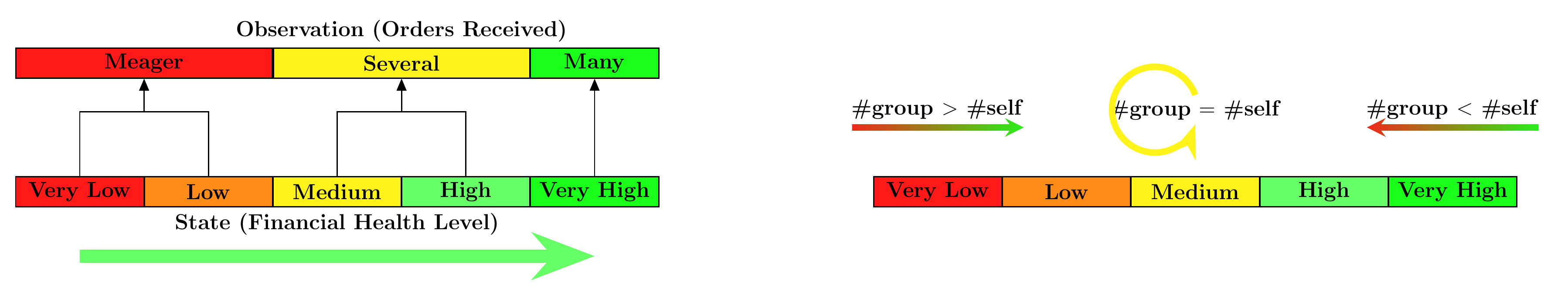}
\caption{\small States, observations, and transition dynamics of the \Org{} domain. The exact specification of the domain is given in the Appendix available in the supplement.}
\label{fig:state}
\end{figure}

The \Org{} domain~\cite{ia2c} is a partially observable multi-agent domain, modeling a typical business organization that features a mix of individual competition with cooperation to improve the financial health of the organization. There are 5 (hidden) states ($s\in S_f$) corresponding to various levels of financial health, which map to 3 observations ($o\in O_f$) that an agent can receive, as shown in Fig.~\ref{fig:state} (left). An agent 0 has 3 action choices ($a_0\in A_0$), viz., {\sf self} (self-interest), {\sf group} (cooperative), and {\sf balance}, the latter benefiting both the group and the individual. Agent 0's reward is comprised of an individual (competitive) component, $R_0$ that depends on the agent's action ($a_0$), and a group (cooperative) component, $R_G$ that depends on the joint action ($\mathbf{a}$). Let,
\begin{equation}
R_0^t\leftarrow R_0(s^t,a_0^t), \ \ R_G^t\leftarrow R_G(s^t,\mathbf{a}^t)
\label{eqn:R1}
\end{equation}
A key feature of this domain is an additional {\em history-dependent} reward component, $R_{-1}$, that models a bonus payoff based on the organization's previous year performance, specifically a fraction $\phi$ of the previous reward, given by 
\begin{align}
R_{-1}^t = \phi(\sum\nolimits_iR_i^{t-1}+R_G^{t-1}).
\label{eqn:R2}
\end{align}
Joint actions are determined by the {\em number} of agents picking self actions compared to that picking group actions, and this affects the state transitions as shown in Fig.~\ref{fig:state} (right). On the one hand, {\sf self} action yields a higher individual reward to an agent than {\sf balance} and {\sf group} actions, it also damages the financial health of the organization if too many agents act in a self-interested manner. On the other hand, {\sf group} action improves the financial health at the expense of individual reward. Thus, with the objective of optimizing $\mathbb{E}_{trajectories}\left [\sum_t \gamma^t(R_G^t + R_i^t + R_{-1}^t)\right ]$, agent $i$ needs to balance greed with group welfare to optimize long-term payoff. He, Banerjee, and Doshi~\cite{ia2c} presents the optimal policy where all agents pick {\sf self} when $observation = many$, and {\sf group} when $observation = meager$, but $\#group=\#self+1$ when $observation=several$. It also presents an I-POMDP formulation of this domain, where the history-dependent reward is explicitly made a (additional) feature of the state to preserve the Markov property. This is elaborated in the next subsection. 

\subsection{Interactive A2C for RL in mixed settings}
\label{subsec:IA2C}

Interactive advantage actor critic (IA2C)~\cite{ia2c} is a decentralized actor-critic method designed for egocentric RL in interactive partially observable Markov decision processes (I-POMDP). I-POMDPs are a generalization of POMDPs~\cite{pomdp} to sequential decision-making in multi-agent environments~\cite{ipomdp,Doshi12:Decision}. We first discuss the I-POMDP formulation of \Org{}, followed by the IA2C algorithm.

In order to capture the history-dependent reward in \Org{}, an extra state feature $S_r$ is introduced. An I-POMDP for agent $0$ in this environment with $N$ other agents is defined as,
\[
\text{I-POMDP}_0 = \langle IS_0, A, T_0, O_0, Z_0, R_0, OC_0 \rangle
\]
\noindent $\bullet~IS_0$ denotes the interactive state space, $IS_0=S_f\times S_r\times\prod_{j=1}^N M_j$. This includes the physical (financial) state $S_f$, and the previous-step reward as an additional state feature $S_r$, as well as models of the other agent $M_j$, which may be intentional (ascribing beliefs, capabilities and preferences) or subintentional~\cite{intention}. Examples of the latter are probability distributions and finite state machines. In this paper, we ascribe subintentional models to the other agents, $m_j=\langle \pi_j,h_j\rangle$, $m_j \in M_j$, where $\pi_j$ is $j$'s policy and $h_j$ is its action-observation history.\\
$\bullet~A = A_0 \times \prod_{j=1}^N A_j$ is the set of joint actions of all agents. Let $\mathbf{a}_{-0}$ denote the joint actions of $N$ other agents, $\mathbf{a}_{-0}\in \prod_{j=1}^N A_j$\\
$\bullet~R_0$ defines the reward function for agent $0$, 
\begin{align}
    R_0(\langle s_f,s_r\rangle,a_0,\mathbf{a}_{-0}) = {\cal R}_0(s_f, a_0, \mathbf{a}_{-0}) + \phi \cdot s_r.
\end{align}
${\cal R}_0(s_f, a_0, \mathbf{a}_{-0})$ is the sum of agent $0$'s current individual and group rewards from \Org{}.\\
$\bullet~T_0$ represents the transition function, 
\begin{align}
  T_0(\langle s_f,s_r\rangle, a_0, \mathbf{a}_{-0}, \langle s_f',s_r'\rangle) 
  = \left\{ \begin{array}{ll} T(s_f, a_0, \mathbf{a}_{-0}, s_f'), & \text{if }s_r'={\cal R}_0(s_f, a_0, \mathbf{a}_{-0}) +\phi\cdot s_r\\
    0 & \text{otherwise}\end{array}\right.
\end{align}
where $T$ is the \Org{}'s transition function. The transition function is defined over the physical states and excludes the other agent's models. This is a consequence of the model non-manipulability assumption, which states that an agent's actions do not directly influence the other agent's models.\\
$\bullet~\Omega_0$ is the set of agent $0$'s {\em private} observations.\\ 
$\bullet~W_0: A \times \Omega_0 \rightarrow  [0,1]$ is the private observation function.\\
$\bullet~O_0 = O_f \times O_r$ is the set of agent $0$'s {\em public} observations, where $O_f$ informs about the physical state and $O_r=S_r$, allowing the agent to observe the past reward.\\ 
$\bullet~Z_0$ is the observation function, 
\begin{align}
  &Z_0(a_0, \mathbf{a}_{-0}, \langle s_f, s_r\rangle, \langle s_f',s_r'\rangle, \langle o_f',o_r'\rangle)
  \nonumber\\
  &= \small \left\{ \begin{array}{ll} Z(a_0, \mathbf{a}_{-0}, s_f, s_f', o_f'), & \text{if }(s_r'={\cal R}_0(s_f,a_0, \mathbf{a}_{-0}) +\phi\cdot
    s_r) \land (o_r'=s_r')\\
    0 & \text{otherwise}\end{array}\right.
\end{align} 
The observation function is defined over the physical state space only as a consequence of the model non-observability assumption, which states that other's model parameters may not be observed directly.\\
$\bullet~OC_0$ is the subject agent's optimality criterion, which may be a finite horizon $H$ or a discounted infinite horizon where the discount factor $\gamma \in (0,1)$. 

The subject agent's belief is a distribution over the interactive state space, $b_0\in\Delta(S_f\times S_r\times \prod_{j=1}^N M_j)$. However, this can be factorized as 
\begin{equation}
    b_0(\langle s_f,s_r\rangle,m_1,\ldots, m_N)=b_0(\langle s_f,s_r\rangle)~b_0(m_1|\langle s_f,s_r\rangle)\ldots b_0(m_N|\langle s_f,s_r\rangle)
    \label{eqn:belief-factorization}
\end{equation}
under the assumption that the other agents' models are conditionally independent given the state. Let $\mathbf{a}_{-0}=a_1\ldots,a_N$. Given the agent's prior belief $b_0$, action $a_0$, as well as its public and private observations $o_0'$, $\omega_0'$, the agent updates its belief over agent $j$'s model for $m_j'=\langle \pi_j',h_j'\rangle$ as
\begin{align}
    b_0'(m_j'|b_0,a_0,o_0',\omega_0')\propto & \sum_{\mathbf{a}_{-0}} 
    \left ( \prod_{k=1}^N\sum_{m_k \in M_k} b_0(m_k)~Pr(a_k|m_k) \right )~W_0(a_0,\mathbf{a}_{-0},\omega_0')~\nonumber\\
    & \times \delta_K(\pi_j',\pi_j)~\delta_K(APPEND(h_j,\langle a_j,o_f' \rangle),h_j')\label{eq:model-bu}
\end{align}
where $\delta_K$ is the Kronecker delta function and APPEND returns a string with the second argument appended to its first.  

In IA2C, each agent has its own critic and actor neural network, the former mapping individual observations to joint action values in terms of the agent's own reward function, $Q_0(\langle o_f,o_r\rangle,a_0,\mathbf{a}_{-0})$, and the latter mapping individual observations to individual action probabilities, $\pi_{0,\boldsymbol{\theta}}(a_0|\langle o_f,o_r\rangle)$, $\boldsymbol{\theta}$ is its set of parameters. IA2C estimates advantages as
\[A_0(\langle o_f,o_r\rangle,a_0,\hat{\mathbf{a}}_{-0}) = avg\left [r + \gamma Q_0(\langle o_f', o_r'\rangle,a_0',\hat{\mathbf{a}}_{-0}') - Q_0(\langle o_f,o_r\rangle,a_0,\hat{\mathbf{a}}_{-0})\right ]\]
while the actor's gradient is estimated as 
\[avg[\nabla_{\boldsymbol{\theta}}\log\pi_{0,\boldsymbol{\theta}}(a_0|\langle o_f,o_r\rangle)~A_0(\langle o_f,o_r\rangle,a_0,\hat{\mathbf{a}}_{-0})]\]
where $r,\langle o_f',o_r'\rangle$ and $a_0'$ are samples, $\hat{\mathbf{a}}_{-0}$ and $\hat{\mathbf{a}}_{-0}'$ are {\em predicted} actions, and the $avg$ is taken over sampled trajectories. In contrast with previous multi-agent deep RL algorithms, IA2C does not require direct exchange of information among RL agents, such as actions and/or gradients. Rather, agents predict each others' actions using their dynamic beliefs over models, updated using their noisy private observations via a belief filter integrated into the critic network to constitute joint actions. Agent $0$'s belief about agent $j$'s model $m_j' = (\pi_j', h_j')$ is updated based on its prior belief over $m_j = (\pi_j, h_j)$, as given in Eq.~\ref{eq:model-bu}.

\section{Many-agent reinforcement learning}
\label{sec:mang-agent}

The primary challenge in scaling multi-agent RL to many agents is the exponential growth of the joint action space. However, if the population is {\em homogeneous} in that all the agents have the same action space ($A_0 = A_1 = \ldots = A_N$)  and the domain exhibits the {\em action anonymity} property, which means that both the dynamics and the rewards depend on the count distribution of actions in the population, while not needing the agents' identities, then we may potentially scale.

\subsection{Mean field approximation}

Under the conditions of population homogeneity and action anonymity, Yang et al.~\cite{MF}, inspired by the mean field theory~\cite{MFT}, utilizes the {\em mean action} of the immediate neighborhood in place of the global joint action. Note that an action can be represented as a one-hot encoding with each component indicating one of the possible actions: $a = [a^1, ..., a^{|A|}]$. The mean action $\bar{a}_0$ is calculated based on the neighborhood of agent 0 as:
\begin{align*}
    \bar{a}_0 = \frac{1}{Ng_0} \sum\nolimits_{j \in Ng(0)} a_j
\end{align*}
where agent $j$ is a neighbor of 0 and $Ng(0)$ is the index set of the neighboring agents of $0$ with size $Ng_0 = |Ng(0)|$. The one-hot action for each neighbor $j$ can be expressed in terms of the sum of $\bar{a}_0$ and a small fluctuation $\Delta a_{0,j}$ as, $a_j = \bar{a}_0 + \Delta a_{0,j}$.

Importantly, the replacement of the other agents' joint action by the mean action is made possible by factorizing the joint action Q-function for agent $i$ using only the pairwise local interactions: 
\begin{align}
Q_0(s, a_0, \mathbf{a}_{-0}) & \approx \frac{1}{Ng_0}\sum\nolimits_{j \in Ng(0)} Q_0(s, a_0, a_j)
\approx Q_0(s, a_0, \bar{a}_0).
\label{eq:MF}
\end{align}
Yang et al.~\cite{MF} notes that the second approximation in~\eqref{eq:MF} is not significant under the conditions of a twice-smooth Q-function and the previously mentioned population homogeneity. 

\begin{wrapfigure}{r}{1.5in} 
\centerline{\includegraphics[width=1.25in]{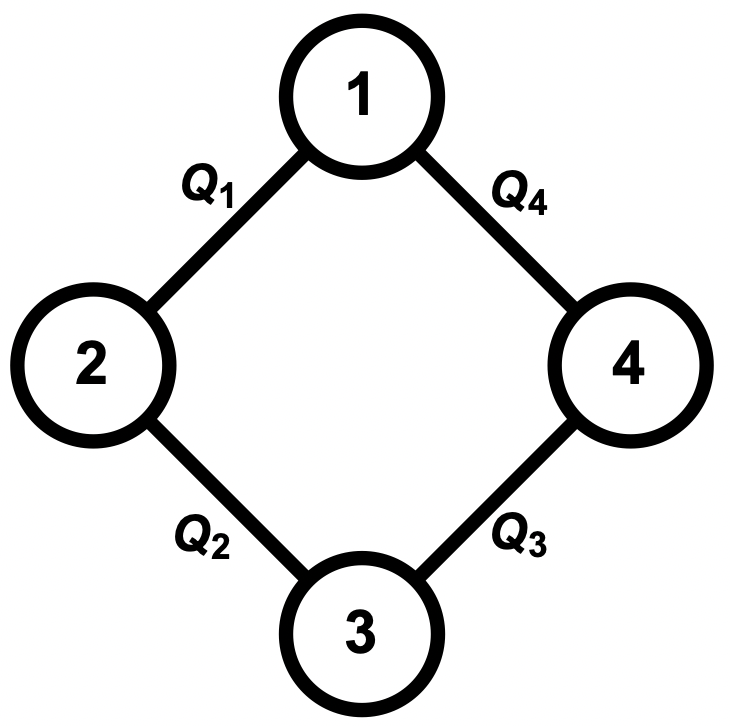}}
\caption{\small The \Org{} domain with 4 worker-agents exhibiting a diamond interaction graph.}
\label{fig:diamond}
\end{wrapfigure}

However, we show that the pairwise interaction basis for the mean-field approximation could be a source of significant error. To demonstrate this, consider an instance of the \Org{} domain in which the worker-agent interactions take the shape of a diamond topology, as shown in Fig.~\ref{fig:diamond}. For illustration, we consider a single-shot interaction between the agents. The topology allows us to factorize the joint action Q-function, which can be written as:
\begin{align}
    Q(a_1,a_2,a_3,a_4) =  Q_1(a_1,a_2) + Q_2(a_2,a_3) + Q_3(a_3,a_4) 
    + Q_4(a_1,a_4).
\label{eq:potential}
\end{align}

Next, we may use Guestrin, Koller, and Parr's well-known value factorization technique~\cite{factor} to obtain the joint action for the agents which optimizes Eq.~\ref{eq:potential} given these potentials. Briefly, optimizing out agent 4's action $a_4$ yields $e_4(a_1,a_3) = \max_{a_4}[Q_3(a_3,a_4) + Q_4(a_1,a_4)]$. Then, $a_3$ can be optimized out as $e_3(a_1,a_2) = \max_{a_3}[Q_2(a_2,a_3) + e_4(a_1,a_3)]$ followed by optimizing out $a_2$: 
$e_2(a_1) = \max_{a_2}[Q_1(a_1,a_2) + e_3(a_1,a_2)]$. The optimal action for agent 1 is then  $\arg\max_{a_1}e_2(a_1)$. Subsequently, we may obtain the optimal actions for the other three agents from $e_2$, $e_3$, and $e_4$.

Recall the reward function of an individual agent in the \Org{} domain from Eqs.~\ref{eqn:R1},~\ref{eqn:R2} in Section~\ref{subsec:org}. Then, $Q_1(a_1,a_2)$ is the potential function of the interaction between agents 1 and 2. Equation~\ref{eq:MF} suggests that $Q_1$ be obtained as, $Q_1(a_1,a_2)  = R_1(a_1) + R_2(a_2) + R_G(a_1,a_2)$; here, the group component $R_G$ of the reward function is limited to the pair of agents participating in the interaction. The remaining Q-potentials on the right-hand side of Eq.~\ref{eq:potential} are obtained similarly.  As we show in the Appendix, the value factorization procedure now yields two optimal joint actions $\langle${\sf self}, {\sf group}, {\sf group}, {\sf group}$\rangle$ and $\langle${\sf group}, {\sf self}, {\sf self}, {\sf group}$\rangle$ both yielding a total reward of 12.~\footnote{Alternately, Eq.~\ref{eq:MF} suggests that we obtain an agent-centric reward using the mean action of its neighborhood to represent others. In other words, for agent 1 obtain $Q_1(a_1,\bar{a}_1) = R_1(a_1) + R_G(a_1,\bar{a}_1)$, and analogously for the other agents. This allows us to construct a normal-form game with the egocentric Q-values contingent on the joint actions as the payoffs for each agent. A search for the Nash equilibrium of this game yields the joint action $\langle${\sf self}, {\sf group}, {\sf self}, {\sf group}$\rangle$ yielding a total reward of 12.}  

But, the \Org{} actually does better and the above joint actions are not the optimal solution. Relaxing the pairwise interaction constraint, let us obtain the potential $Q_1$ as: $Q_1(a_1,a_2) = R_1(a_1) + R_2(a_2) + R_G(a_1,a_2,a_3,a_4)$. Notice that the group component of the reward is relaxed to include all agents in the organization. Applying Guestrin, Koller, and Parr's value factorization technique as before involves optimizing out $a_4$: $e_4(a_1,a_3) = \max_{a_4}[Q_3(a_3,a_4) + Q_4(a_1,a_4)]$. Notice that action $a_2$ appearing in the group component $R_G(a_1,a_2,a_3,a_4)$ included in both $Q_3$ and $Q_4$ remains unspecified and we set its value as the one that maximizes $Q_3(a_3,a_4) + Q_4(a_1,a_4)$. Similarly, in the next step of optimizing out $a_3$ as $e_3(a_1,a_2) = \max_{a_3}[Q_2(a_2,a_3) + e_4(a_1,a_3)]$, the unspecified action $a_4$ included in the group component of $Q_2(a_2,a_3)$ is the one that maximizes $Q_2$. Moving forward, we optimize out $a_2$ as, $e_2(a_1) = \max_{a_2}[Q_1(a_1,a_2) + e_3(a_1,a_2)]$, while choosing $a_3$ and $a_4$ in $Q_1(a_1,a_2)$, which maximize the $Q_1$ potential, leading to the final: $\arg\max_{a_1}e_2(a_1)$. All other agents' actions are obtained analogously. This procedure yields the globally optimal joint action $\langle${\sf self}, {\sf balance}, {\sf group}, {\sf group}$\rangle$ and a total reward of 13.

\subsection{Many-agent RL using action configurations}
\label{subsec:configuration}

The mean field action offers a scalable way to model agent populations under the conditions of population homogeneity and action anonymity. But, its use is based on value approximations as we demonstrated in the previous section; our experiments reveal that this approximation can be costly. We present an alternate way to model agent populations that is both scalable and lossless under the same population conditions, and which has been effective in scaling decision-theoretic planning.

\subsubsection{Action configurations}

We begin by defining the concept of a configuration and characterize its properties.

\begin{defn}[Configuration] Define a configuration denoted by $\C{}$ as a vector of counts of the distinct actions performed by the agents, $\C{}^\mathbf{a}{} = \langle \#a^1, \#a^2, \ldots, \#a^{|A|} \rangle$, where $\#a^1$ denotes the count of an action $a^1$ in the joint action $\mathbf{a}$. Denote by $\boldsymbol{\C}$ the set of all configurations.     
\end{defn}

\begin{defn}[Projection] Define a projection function $\delta$ as a mapping $\delta: A \rightarrow \boldsymbol{\C}$, which maps a joint action to its corresponding configuration. 
\end{defn}

For example, $\delta$ projects the joint action $\mathbf{a} = \langle$ {\sf self}, {\sf self}, {\sf group}, {\sf group} $\rangle$ in the \Org{} domain instance of Fig.~\ref{fig:diamond} to the configuration vector, $\C{}^\mathbf{a}= \langle 2, 2, 0 \rangle$, where the first component of the vector gives the count of action {\sf self}, the second gives the count of {\sf group}, while the third component gives the count of the {\sf balance} action.

Observe that $\delta$ is a many-one mapping as multiple distinct joint actions, which are permutations of each other, yield the same configuration vector. In other words, for any $s_f$, $s_r$, $a_0$, $s_f'$, $s_r'$, $\mathbf{a}_{-0}$, and a permutation of $\mathbf{a}_{-0}$ denoted as $\mathbf{\dot{a}}_{-0}$, we have:

\centerline{$T_0(\langle s_f, s_r \rangle, a_0, \mathbf{a_{-0}},\langle s_f', s_r' \rangle) = T_0(\langle s_f, s_r \rangle, a_0, \mathbf{\dot{a}_{-0}},\langle s_f', s_r' \rangle) = T_0(\langle s_f, s_r \rangle, a_0, \C^\mathbf{a_{-0}},\langle s_f', s_r' \rangle)$,} 
\centerline{$Z_0(a_0, \mathbf{a_{-0}}, \langle s_f, s_r \rangle, \langle s_f', s_r' \rangle, \langle o_f', o_r' \rangle) = Z_0(a_0, \mathbf{\dot{a}_{-0}}, \langle s_f, s_r \rangle, \langle s_f', s_r' \rangle, \langle o_f', o_r' \rangle)$} \centerline{$= Z_0(a_0, \C^\mathbf{a_{-0}},\langle s_f, s_r \rangle, \langle s_f', s_r' \rangle, \langle o_f', o_r' \rangle)$,} 
\centerline{$W_0(a_0, \mathbf{a_{-0}}, \omega_o') = W_0(a_0, \mathbf{\dot{a}_{-0}},\omega_o') = W_0(a_0, \C^\mathbf{a_{-0}},\omega_o')$,~~~ \textit{and}} 
\centerline{$R_0(\langle s_f, s_r \rangle, a_0, \mathbf{a_{-0}}) = R_0(\langle s_f, s_r \rangle, a_0, \mathbf{\dot{a}_{-0}}) = R_0(\langle s_f, s_r \rangle, a_0, \C^\mathbf{a_{-0}})$}
where $\delta(\mathbf{a_{-0}})=\delta(\mathbf{\dot{a}_{-0}})=\C^\mathbf{a_{-0}}$. As such, we may not recover the original joint action back from the configuration -- a direct consequence of the action anonymity property. The above equivalences due to permutation invariance in the environment's dynamics and the agent's observation capabilities and preferences naturally lead to the following property of the Q-function:

\centerline{$Q_0(\langle o_f, o_r \rangle,a_0,\mathbf{a}_{-0}) = Q_0(\langle o_f, o_r \rangle ,a_0,\mathbf{\dot{a}}_{-0}) = Q_0(\langle o_f, o_r \rangle,a_0,\C^\mathbf{a_{-0}})$.}

Subsequently, the advantage function $A_0(\langle o_f, o_r \rangle,a_0,\mathbf{a_{-0}})$ is also rewritten with the projection to the configuration. {\em A key advantage of using configurations is that the space of vectors of action counts is polynomial in the number of agents in comparison to the exponential growth of the joint action space as the number of agents grows.}

We adapt the IA2C method of Section~\ref{subsec:IA2C} to include action configurations to enable many-agent RL in partially-observable settings, and label this new method as IA2C$^{++}$. IA2C's belief filter is modified and a new dynamic programming module is prepended to the belief filter in the critic.

\begin{algorithm} [!t]
\caption{\small Computing configuration distribution $Pr(\C|b_0(M_1), b_0(M_2), \ldots, b_0(M_N))$}
\label{alg}
\small
\begin{algorithmic}
\REQUIRE $\langle b_0(M_1), b_0(M_2), \ldots, b_0(M_N) \rangle$ 
\ENSURE $P_N$, which is the distribution $Pr(\boldsymbol{\C}^\mathbf{a_{-0}})$ represented as a trie.
\STATE Initialize $c^{a_i}_0 \leftarrow (0,\dots,0)$, $P_0[c^{a_i}_0] \leftarrow 1.0$
\FOR {$k = 1$ to $N$}
\STATE Initialize $P_k$ to be an empty trie
\FOR {$c^{a_i}_{k-1}$ from $P_{k-1}$}
\FOR {$a^{a_i}_k \in A^{a_i}_k$ such that $\pi^{a_i}_k(a^{a_i}_k) > 0$}
\STATE $c^{a_i}_k \leftarrow c^{a_i}_{k-1}$
\IF {$a^{a_i}_k \neq \emptyset$}
\STATE $c^{a_i}_k(a^{a_i}_k) \overset{+}{\leftarrow} 1$
\ENDIF
\IF {$P_k[c^{a_i}_k]$ does not exist}
\STATE $P_k[c^{a_i}_k] \leftarrow 0$
\ENDIF
\STATE $P_k[c^{a_i}_k] \overset{+}{\leftarrow} P_{k-1}[c^{a_i}_{k-1}] \times \pi^{a_i}_k(a^{a_i}_k)$
\ENDFOR
\ENDFOR
\ENDFOR
\RETURN $P_N$
\end{algorithmic}
\label{alg:dp}
\end{algorithm}

\subsubsection{Belief update with configurations}

Equation~\ref{eq:model-bu} in Section~\ref{subsec:IA2C} gives the update of agent 0's belief over {\em one} other agent's possible models $M_j$, and this is performed for each other agent $j= \{1, 2, \ldots, N\}$ -- growing linearly in $N$. Joint action in the private observation function $W_0$ is now replaced by the configuration, as introduced previously. But, this also necessitates an additional term in the equation as we show below.
\begin{small}
\begin{align}
b_0'(m_j'|b_0,a_0,o_0',\omega_0')\propto & \sum\limits_{m_j \in M_j} b_0(m_j) \sum\limits_{a_j} Pr(a_j|m_j) ~\sum\limits_{\C \in \boldsymbol{\C}^\mathbf{a_{-0}}} Pr(\C|b_0(M_1), b_0(M_2),\nonumber\\ 
&~~ \ldots, b_0(M_N))~W_0(a_0,\C,\omega_0')~\delta_K(\pi_j,\pi_j')\delta_K(APPEND(h_j, \langle a_j,o_f' \rangle),h_j').
\label{eq:model-bu-cfg}
\end{align}
\end{small}
Here, $Pr(\C|b_0(M_1), b_0(M_2), \ldots, b_0(M_N))$ is the probability of a configuration in the distribution over the set of configurations $\boldsymbol{\C}^\mathbf{a_{-0}}$. The distribution is obtained from agent 0's factored beliefs over the models of each other agent using a known dynamic programming procedure~\cite{configuration} that is outlined in Algorithm~\ref{alg:dp}. The algorithm takes as input just $N$ beliefs each of size $|M_j|$ compared to a single large belief of exponential size $|M_j|^N$, which is a benefit of the belief factorization shown in Section~\ref{subsec:IA2C}.

\subsubsection{IA2C$^{++}$ architecture}

\begin{figure}[t!]
\centerline{\includegraphics[width=4.5in]{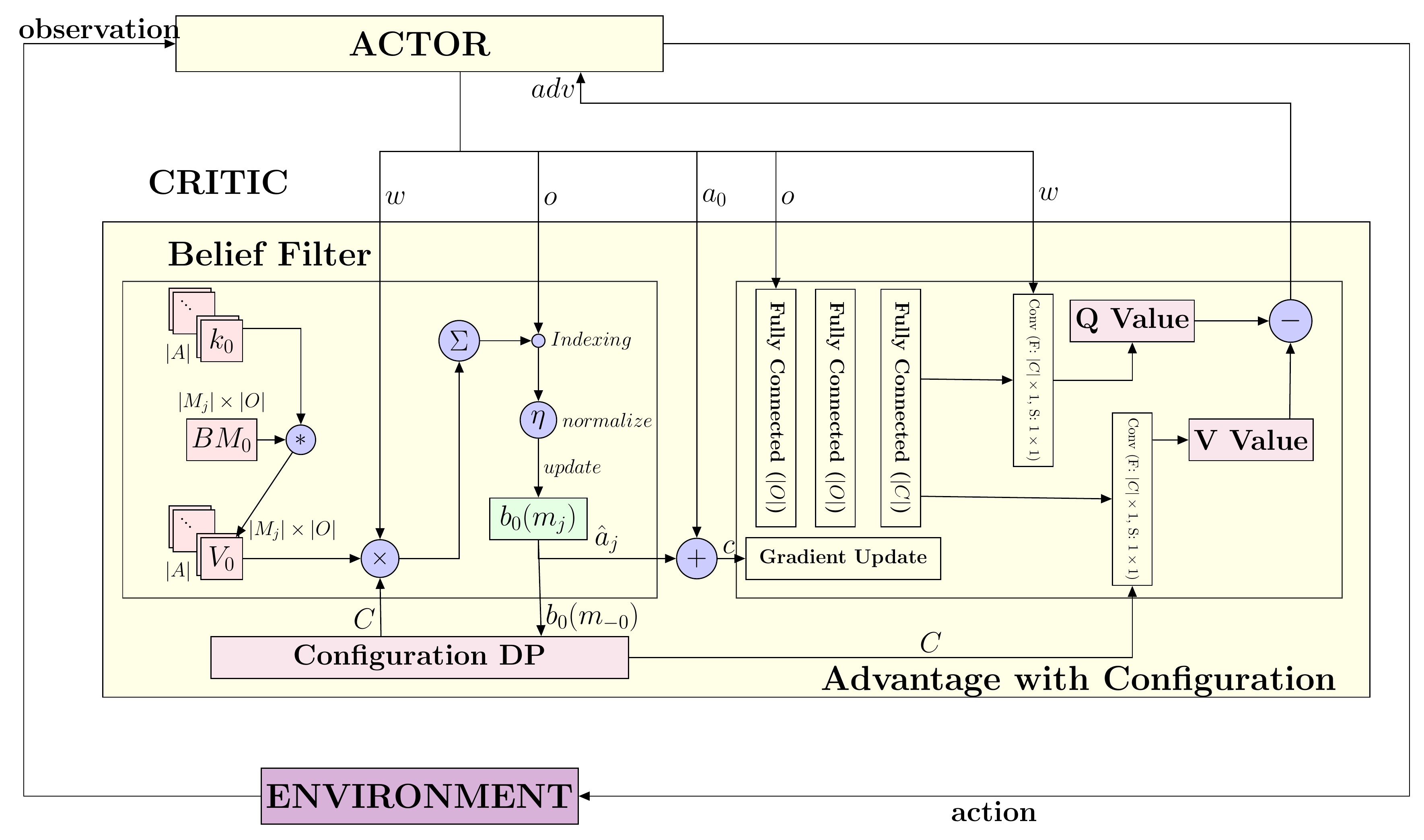}}
\caption{\small The belief filter in the critic utilizes the distribution over configurations computed using dynamic programming to update the agent's belief over models. Predicted actions are projected to their corresponding configurations and used in obtaining the advantage value.}
\label{fig:network}
\vspace{-0.05in}
\end{figure}

We illustrate IA2C$^{++}$'s architecture in the schematic of Fig.~\ref{fig:network}. It consists of two main components: the actor and the critic. The expression in~\eqref{eq:gradient} below gives the actor's revised gradient, which is  updated as the subject agent 0 interacts with the environment:
\begin{align}
    avg \left [ \nabla_{\boldsymbol{\theta}} ~log ~\pi_{0,\boldsymbol{\theta}}(a_0|\langle o_f, o_r \rangle) ~A_0(\langle o_f, o_r \rangle, a_0,\C^\mathbf{a_{-0}}) \right ].
\label{eq:gradient}
\end{align}
Notice that the actions of the other agents traditionally appearing in the advantage function are now replaced with its projected configuration $\C^\mathbf{a_{-0}} (= \delta(\mathbf{a_{-0}}))$. This new advantage function is computed by the critic as:
\begin{align*}
    A_0(\langle o_f, o_r \rangle, a_0, \C^\mathbf{a_{-0}}) = avg \left [r + \gamma ~Q_0(\langle o_f', o_r' \rangle, a_0', \C^\mathbf{a_{-0}'}) - Q_0(\langle o_f, o_r \rangle, a_0, \C^\mathbf{a_{-0}}) \right ]
\end{align*}
where $r$, $\langle o_f', o_r' \rangle$, and $a_0'$ are samples, $\mathbf{a_{-0}}$ and $\mathbf{a_{-0}'}$ are the {\em predicted} most-likely joint actions of the other agents for the current and next step, respectively, replaced by their corresponding configurations, and $avg$ is taken over the sampled trajectories. An agent $j$'s predicted action for the next time step is obtained by first sampling its model from the updated $b_0'(m_j')$, where the update occurs as per Eq.~\ref{eq:model-bu-cfg}. The sampled model yields an action distribution from which the action is sampled. This procedure is performed for each other agent and the corresponding configuration is obtained. 

Thus, the actor network forwards the public and private observations from the environment to the belief filter in the critic. The belief filter first runs the dynamic programming procedure in Algorithm~\ref{alg:dp}, uses the output distribution over configurations to then update agent 0's belief over other agents' models, and predicts their actions. The projection operator yields the corresponding current and next time-step configurations, all of which is sent to the critic neural net for gradient-based updating and then to the advantage module for computing the advantage function. The latter is sent back to the actor component for its gradient update.

We implement the actor neural network with one input layer for the observations, two hidden layers one with tanh and the other with ReLU activation, followed by the output layer. The critic network consists of one input layer for the observations, one hidden layer with tanh activation followed by the output layer. All layers are fully connected to the next layer.




\section{Experiments}
\label{sec:experiments}

\begin{figure}[!t]
\centerline{
\subfigure[\small Connected]{\label{fig:full}\includegraphics[width=0.16\textwidth]{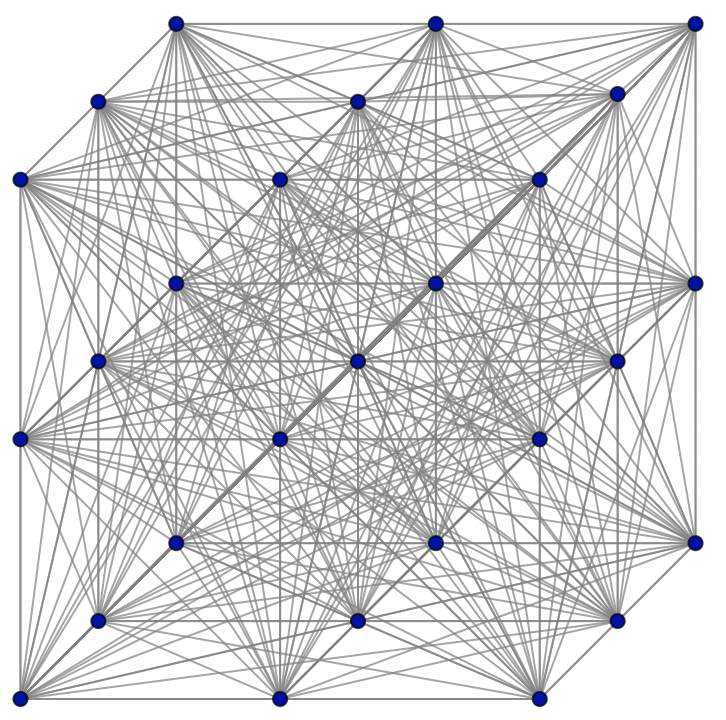}}
\subfigure[\small Tree]{\label{fig:tree}\includegraphics[width=0.32\textwidth]{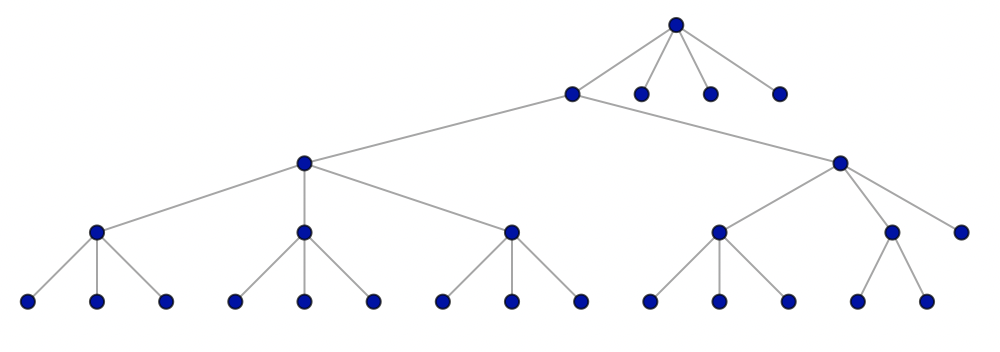}}
\subfigure[\small Lattice]{\label{fig:lattice}\includegraphics[width=0.16\textwidth]{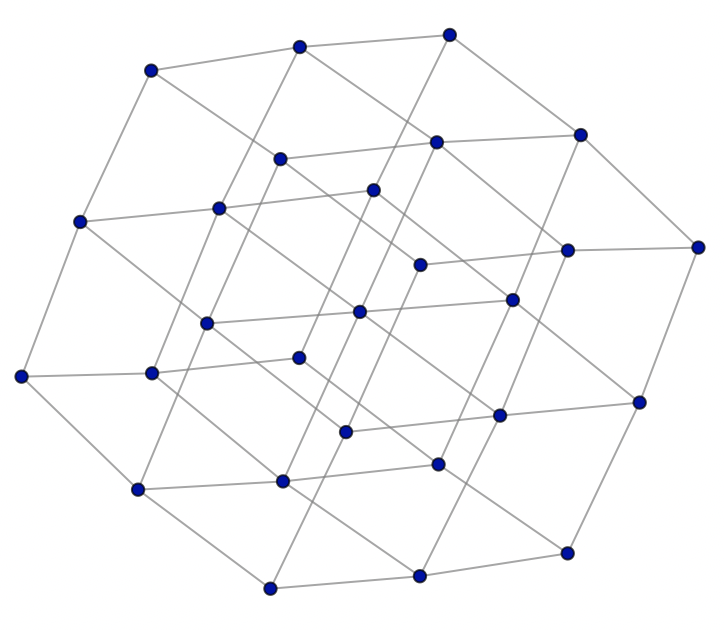}}
\subfigure[\small Circle]{\label{fig:circle}\includegraphics[width=0.16\textwidth]{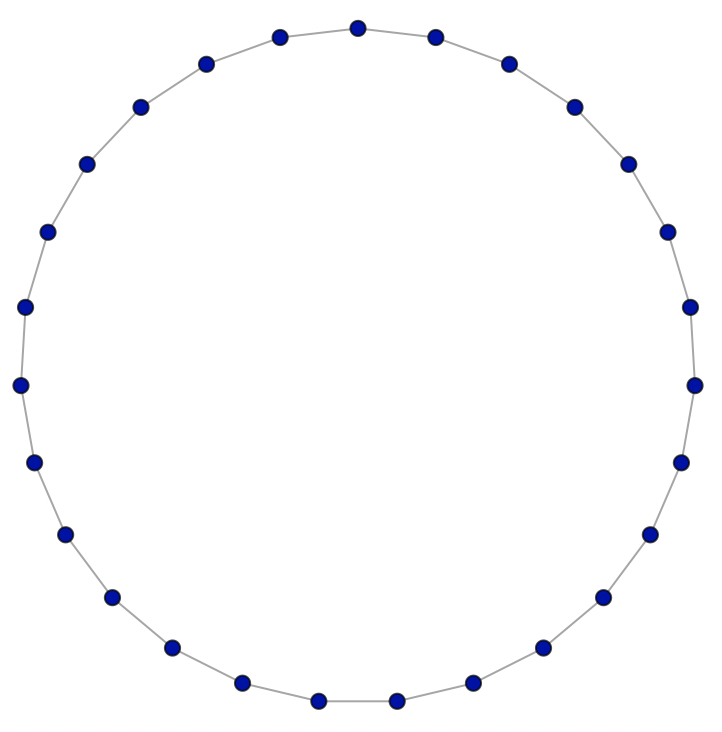}}
\subfigure[\small Star]{\label{fig:star}\includegraphics[width=0.16\textwidth]{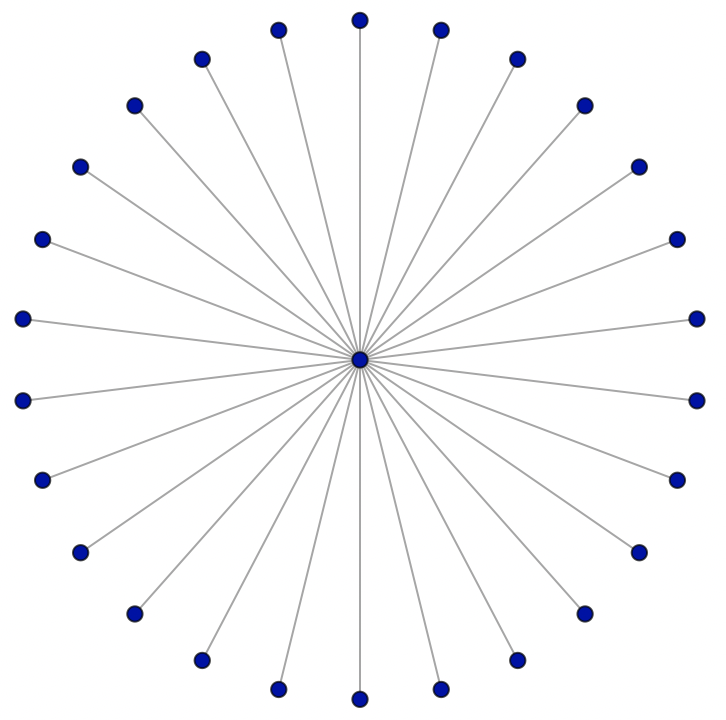}}}
\caption{\small Interaction topologies for the \Org{} domain with 27 agents. We experiment with (a) fully connected, (b) tree, (c) lattice, (d) circle, and (e) star structures. Each dot denotes an agent.}
\vspace{-0.05in}
\end{figure}

We instantiate IA2C and MADDPG~\cite{maddpg} with mean-field approximation and configuration, and label their scalable versions as IA2C$^{++}$(MF), IA2C$^{++}$(CF), MADDPG$^{++}$(MF), and MADDPG$^{++}$(CF), respectively. We conduct experiments using these four methods on \Org{} with five different graph structures of agent connectivity. Our code is available on GitLab and will be publicly released upon publication. The original \Org{} presumes a fully connected structure (Fig.~\ref{fig:full}) between agents, which models a group of employees working on the same project. A tree structure (Fig.~\ref{fig:tree}) models the common hierarchical structure of an organization. We study additional structures that accommodate loopy connections, specifically the lattice (Fig.~\ref{fig:lattice}) and circle topologies (Fig.~\ref{fig:circle}). Furthermore, from a macro perspective, we can model an organization with its subsidiary units using a star structure (Fig.~\ref{fig:star}). 

The five structures differ by the number of neighborhoods and number of agents in each neighborhood. For example, in the fully connected structure, all agents share one single neighborhood; on the contrary, each agent forms a neighborhood with its left and right neighbors in the circle structure. We demonstrate that the various neighborhoods indeed impact the performance of the mean-field based methods. It is important to note that as \Org{}'s reward function depends on joint actions (hence the number of agents) but not on the graph structure, the globally optimal value and policy are also dependent on the number of agents and not on the graph structure. Consequently, the varying graph structures do not impact the value and policy learned by the configuration based methods.

Table~\ref{tbl:policyvalue} lists the values of converged policies from the four methods. The two configuration based methods (IA2C$^{++}$(CF) and MADDPG$^{++}$(CF)) always converge to the optimal policy no matter the structure, so we show only one column for them. Furthermore, IA2C$^{++}$(MF) and MADDPG$^{++}$(MF) converge to the same policies for any given structure and number of agents, so we show one column for each structure. In the fully connected structure, IA2C$^{++}$(MF) and MADDPG$^{++}$(MF) also converge to optimal policies, as the pairwise neighborhood basis of the mean field encompasses all other agents and induces no approximation.
However, in the tree, lattice, circle, and star structures, the mean-field instantiations are approximations and do not converge to optimal policies. Moreover, MADDPG$^{++}$(MF) and IA2C$^{++}$(MF) converged to better policies in lattice and circle structure than star and tree structures. This is because star and tree structures contain many neighborhoods of only two agents, making it harder for the mean-field approximation to coordinate across many small neighborhoods without knowing the actions outside the neighborhoods.

\begin{table}[ht!]
\caption{\small Policy value comparison between the configuration and mean-field based methods on the fully-connected, tree, lattice, circle, and star interaction topologies in the \Org{} domain. Both MADDPG$^{++}$ and IA2C$^{++}$ converge to the same policies but differ in run times. While the configuration instantiation consistently converged to the optimal policy in all topologies, mean-field approximation converged to the optimal policy in the fully connected structure only.}
\label{tbl:policyvalue}
\centerline{
\begin{small}
\begin{tabular}{|c||c|c|c|c|c|c|}
\hline
\multirow{2}{*}{\shortstack[l]{\# of \\ agents}}& {Configuration} & \multicolumn{5}{c|}{Mean field approximation} \\
\cline{2-7}
& {\bf All} & \textbf{Full} & \textbf{Tree} & \textbf{Lattice} & \textbf{Circle} & \textbf{Star} \\
\hline
27 & 3,180 & 3,180 & 1,620 & 2,820 & 2,700 & 1,740 \\
\hline
40 & 4,710 & 4,710 & 2,400 & 3,840 & 3,960 & 2,520 \\
\hline
60 & 7,110 & 7,110 & 3,600 & 6,000 & 6,000 & 3,720 \\
\hline
80 & 9,510 & 9,510 & 4,800 & 7,650 & 7,920 & 4,920 \\
\hline
100 & 11,910 & 11,910 & 6,000 & 9,450 & 9,960 & 6,120 \\
\hline
\end{tabular}
\end{small}}
\vspace{-0.05in}
\end{table}

\begin{figure}
\centering
\subfigure[\small MADDPG$^{++}$($\cdot$)]{\label{fig:comp1}\includegraphics[width=0.45\textwidth]{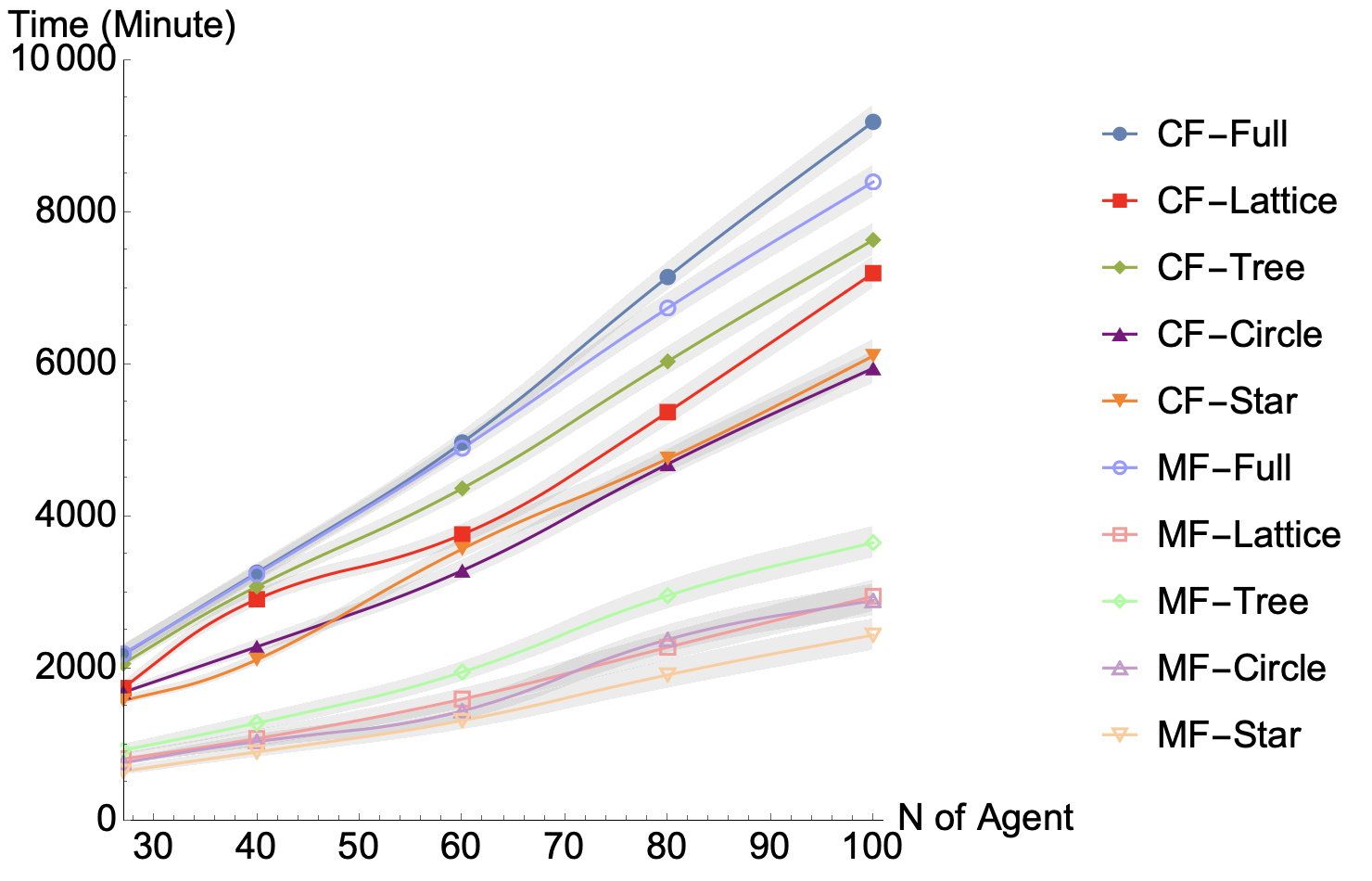}}
\subfigure[\small IA2C$^{++}$($\cdot$)]{\label{fig:comp2}\includegraphics[width=0.45\textwidth]{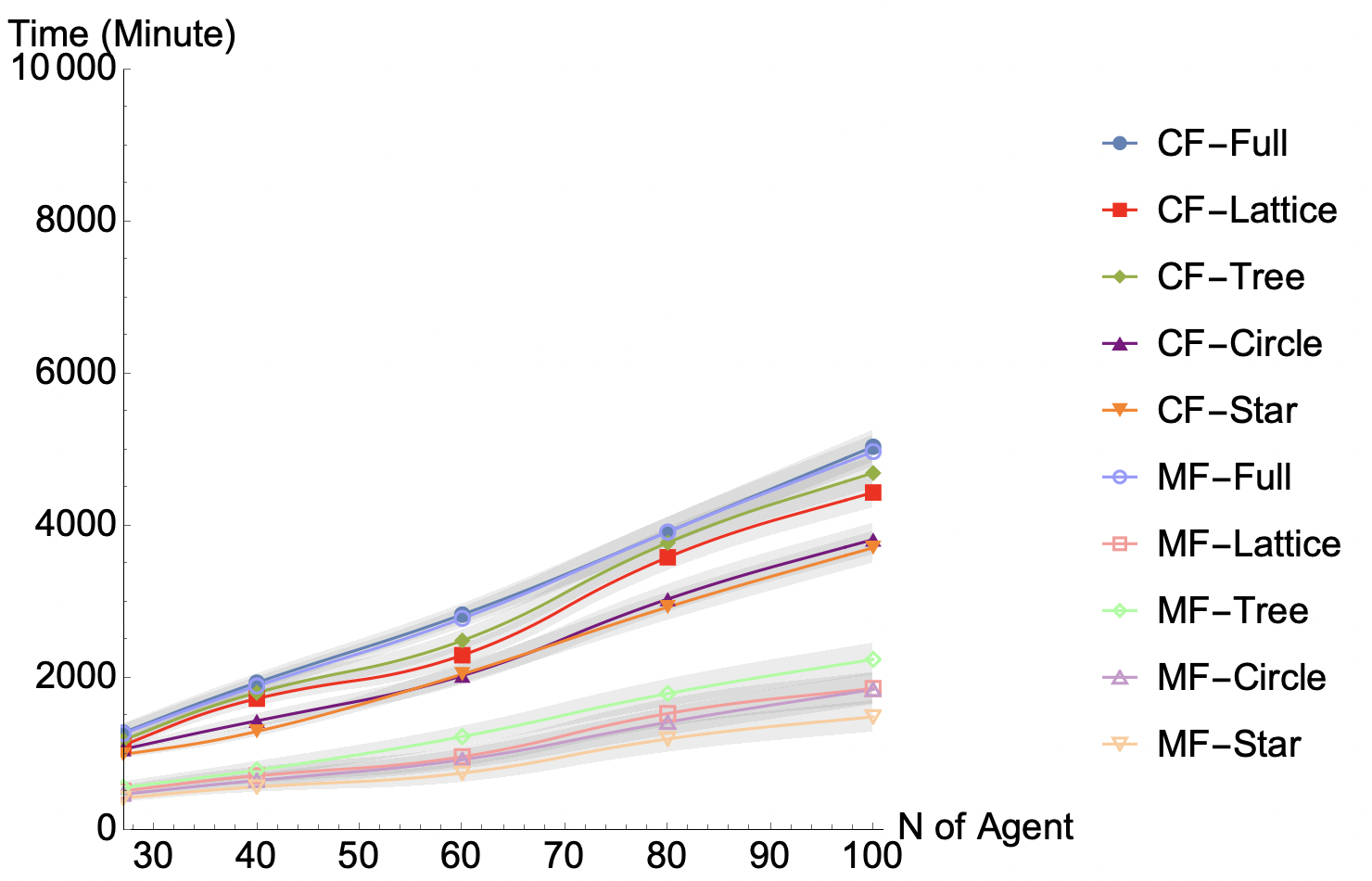}}
\caption{\small Run time comparison between mean-field approximation and configuration instantiations on MADDPG$^{++}$ and IA2C$^{++}$. Run times are the average of three runs. Notice that MADDPG$^{++}$($\cdot$) generally takes more time to converge than IA2C$^{++}$($\cdot$). All experiments are run on a Linux platform with 2.3GHz quad-core i7 processor with 8GB memory. The discount factor is 0.9 and the learning rates for the actor and critic networks are 0.001 and 0.005, respectively. }
\label{fig:timecompare}
\vspace{-0.1in}
\end{figure}

Figure~\ref{fig:timecompare} shows the run time of MADDPG$^{++}$ and IA2C$^{++}$ with mean-field approximation and configuration instantiations. All experiments are run on a Linux platform with 2.3GHz quad-core i7 processor with 8GB memory. We observe that in general the mean-field approximation instantiations converge {\em faster} and require less episodes to train than their configuration based counterparts. On average, the mean-field approximation instantiations require 43.35\% less episodes to converge than the configuration instantiations across all topologies and numbers of agents. This can be viewed as a trade-off against mean field's convergence to suboptimal policies. Among the various interaction topologies, the fully connected and lattice structures consume the most time to converge due to their larger neighborhoods. 
Furthermore, IA2C$^{++}$(MF) and IA2C$^{++}$(CF) (Fig.~\ref{fig:comp2}) are faster to converge than MADDPG$^{++}$(MF) and MADDPG$^{++}$(CF) (Fig.~\ref{fig:comp1}), consistent with the observation in He, Banerjee, and Doshi~\cite{ia2c} that IA2C converges faster than MADDPG. This is a benefit of the belief filter that can predict the other agents' actions more accurately than the maximum likelihood estimation of MADDPG.

\section{Related work}
\label{sec:relatedworks}
\vspace{-0.05in}

Mean-field Q learning (MF-Q)~\cite{MF} is a Q-learning RL algorithm that scales to many agents. If the agents are indistinguishable and independent from each other, they are represented as a single virtual agent who performs a mean action. The Q-value of the state and joint action is approximated by the Q-value of the state and mean action. A mean-field actor-critic (MF-AC) is also presented, but experiments show that MF-AC rarely improves on MF-Q. Both MF-Q and MF-AC are tested on domains with only one neighborhood or with several neighborhoods that are nearly fully isolated from each other. In contrast, we evaluate MADDPG$^{++}$ and IA2C$^{++}$ on domains with multiple neighborhoods that are connected to each other. 

In contrast to our use of action configurations, Verma, Varakantham, and Lau~\cite{de} utilize state configurations under anonymity in the form of counts of agents located in various zones (the agents are taxis). Other agents' actions in the transition and reward functions are now replaced by state configurations without loss of information in the taxi domain. Algorithms that extend deep Q-networks and A2C to these settings are presented. However, the state is perfectly observed obviating the need for distributions over configurations and the  experiments were mostly limited to just 20 agents operating in about  100 zones.       


\section{Concluding remarks}
\label{sec:conclusion}
\vspace{-0.05in}

We presented new scalable instantiations of existing actor-critic MARL algorithms built on the notion of action anonymity that allows joint actions to be replaced by action configurations. Owing to a polynomial-time dynamic programming approach for computing distributions over configurations based on model beliefs, our instantiations of recent algorithms -- IA2C and MADDPG -- are able to scale polynomially with the number of agents, unlike the original algorithms. When compared to a recent scalable alternative for MARL that uses mean-field approximation, we found the latter to be significantly faster than using configurations in the cooperative-competitive \Org{} domain. However, this advantage in run time comes at a debilitating cost. The mean-field approximation was unable to learn optimal policies in all but the fully-connected interaction topology of agents. On the other hand, the configuration based approach learned optimal policies in all of the topology structures. Varying topologies are a natural feature of organizations, and indeed of many MARL domains, therefore the ability of the configuration based approach to learn optimal policies in a topology-independent way while being able to scale sets it apart in the sparse field of scalable MARL algorithms. Our future work involves investigating other mixed cooperative-competitive domains where MARL can help learn optimal behaviors.  

\clearpage
\bibliographystyle{abbrv}  
\bibliography{hdbNeurIPS21}  
\end{document}